\documentclass[11pt,a4paper]{article}
\usepackage[hyperref]{emnlp2018}
\usepackage{times}
\usepackage{latexsym}
\usepackage{graphicx}
\usepackage{amsmath,amssymb}
\usepackage{pifont}
\usepackage{array}
\usepackage{hyperref}
\usepackage{booktabs}
\usepackage[small]{caption}
\usepackage{tablefootnote}
\pdfoutput=1

\usepackage[colorinlistoftodos,textsize=scriptsize]{todonotes}

\usepackage{url}

\newcommand{\namedref}[2]{\hyperref[#2]{#1~\ref*{#2}}}

\newcommand{\sectionref}[1]{\namedref{Section}{sec:#1}}
\newcommand{\tableref}[1]{\namedref{Table}{tab:#1}}
\newcommand{\figureref}[1]{\namedref{Figure}{fig:#1}}
\newcommand{\appendixref}[1]{\namedref{Appendix}{appendix:#1}}

\aclfinalcopy 

\title{Revisiting the Importance of Encoding Logic Rules \\ in Sentiment Classification}

\author{Kalpesh Krishna$^{\spadesuit \clubsuit}$ \And Preethi Jyothi$^\spadesuit$ \\\\ Indian Institute of Technology, Bombay$^\spadesuit$ \\ University of Massachusetts, Amherst$^\clubsuit$ \\ \texttt{\{kalpesh,miyyer\}@cs.umass.edu}\\ \texttt{pjyothi@cse.iitb.ac.in} \And Mohit Iyyer$^\clubsuit$}

\date{}

\begin{document}
\maketitle
\begin{abstract}
We analyze the performance of different sentiment classification models 
on syntactically-complex inputs like \textit{A-but-B} sentences. The first contribution of this analysis addresses reproducible research: to meaningfully compare different models, their accuracies must be averaged over far more random seeds than what has traditionally been reported. With proper averaging in place, we notice that the distillation model described in ~\newcite{hu2016harnessing}, which incorporates explicit logic rules for sentiment classification,  is ineffective. In contrast, using contextualized ELMo embeddings~\cite{PetersELMo2018} instead of logic rules yields significantly better performance. Additionally, we provide analysis and visualizations that demonstrate ELMo's ability to implicitly learn logic rules. Finally, a crowdsourced analysis reveals how ELMo outperforms baseline models even on sentences with ambiguous sentiment labels.
\end{abstract}

\section{Introduction}
In this paper, we explore the effectiveness of methods designed to improve sentiment classification (positive vs. negative) of sentences that contain complex syntactic structures. While simple bag-of-words or lexicon-based methods~\cite{pang2005seeing,wang2012simple,iyyer2015deep} achieve good performance on this task, they are unequipped to deal with syntactic structures that affect sentiment, such as contrastive conjunctions (i.e., sentences of the form \textit{``A-but-B''}) or negations. Neural models that explicitly encode word order~\cite{kim2014convolutional},  syntax~\cite{socher2013recursive,tai2015improved} and semantic features~\cite{li2017initializing} have been proposed with the aim of improving performance on these more complicated sentences. Recently,~\newcite{hu2016harnessing} incorporate logical rules into a neural model and show that these rules increase the model's accuracy on sentences containing contrastive conjunctions, while~\newcite{PetersELMo2018} demonstrate increased overall accuracy on sentiment analysis by initializing a model with representations from a language model trained on millions of sentences.



In this work, we carry out an in-depth study of the effectiveness of the techniques in \newcite{hu2016harnessing} and \newcite{PetersELMo2018} for sentiment classification of complex sentences. Part of our contribution is to identify an important gap in the methodology used in \newcite{hu2016harnessing} for performance measurement, which is addressed by averaging the experiments over several executions. With the averaging in place, we obtain three key findings: (1) the improvements in~\newcite{hu2016harnessing} can almost entirely be attributed to just one of their two proposed mechanisms and are also less pronounced than previously reported; (2) contextualized word embeddings~\cite{PetersELMo2018} incorporate the ``A-but-B'' rules more effectively without explicitly programming for them; and (3) an analysis using crowdsourcing reveals a bigger picture where the errors in the automated systems have a striking correlation with the inherent sentiment-ambiguity in the data.

\section{Logic Rules in Sentiment Classification}
Here we briefly review background from~\newcite{hu2016harnessing} to provide a foundation for our reanalysis in the next section.
We focus on a logic rule for sentences containing an \textit{``A-but-B''} structure (the only rule for which \newcite{hu2016harnessing} provide experimental results). Intuitively, the logic rule for such sentences is that the sentiment associated with the whole sentence should be the same as the sentiment associated with phrase \textit{``B''}.\footnote{The rule is vacuously true if the sentence does not have this structure.}

More formally, let $p_\theta(y|x)$ denote the probability assigned to the label $y\in\{+,-\}$ for an input $x$ by the baseline model using parameters $\theta$. A logic rule is (softly) encoded as a variable $r_\theta(x,y)\in[0,1]$ indicating how well labeling $x$ with $y$ satisfies the rule. For the case of \textit{A-but-B} sentences, 
$r_\theta(x,y)=p_\theta(y|B)$ if $x$ has the structure \textit{A-but-B} (and 1 otherwise).
Next, we discuss the two techniques from~\newcite{hu2016harnessing} for incorporating rules into models: \emph{projection}, which
directly alters a trained model, and \emph{distillation}, which progressively
adjusts the loss function during training. 

\paragraph{Projection.} The first technique
is to \emph{project a trained model into a rule-regularized subspace},
in a fashion similar to~\newcite{ganchev2010posterior}.  More precisely, a
given model $p_\theta$ is projected to a model $q_\theta$ defined by the
optimum value of $q$ in the following optimization problem:%
\footnote{The formulation in \newcite{hu2016harnessing} includes another
hyperparameter $\lambda$ per rule, to control its relative importance; when
there is only one rule, as in our case, this parameter can be absorbed into
$C$.}
\begin{align*}
\min_{q,\xi \ge 0}&~ \mathrm{KL}(q(X,Y) || p_\theta(X,Y)) + C\sum_{x\in X} \xi_x\\
\text{s.t. }&~ (1 - \mathbb{E}_{y\leftarrow q(\cdot|x)}[r_\theta(x,y)]) \leq \xi_x
\end{align*}
Here $q(X,Y)$ denotes the distribution of $(x,y)$ when $x$ is drawn
uniformly from the set $X$ and $y$ is drawn according to $q(\cdot|x)$.

\paragraph{Iterative Rule Knowledge Distillation.}
The second technique is to transfer the domain knowledge
encoded in the logic rules into a neural network's parameters.
Following~\newcite{hinton2015distilling}, a ``student'' model $p_\theta$
can learn from the ``teacher'' model $q_\theta$, by using a loss function
$\pi H(p_\theta, P_\text{true}) + (1 - \pi) H(p_\theta, q_\theta)$
during training, where $P_\text{true}$ denotes the distribution implied by the
ground truth, $H(\cdot, \cdot)$ denotes the cross-entropy function, and
$\pi$ is a hyperparameter. ~\newcite{hu2016harnessing} computes $q_\theta$ after every gradient update
by projecting the current $p_\theta$, as described above.
Note that both mechanisms can be combined: After fully training $p_\theta$
using the iterative distillation process above, the projection step can be
applied one more time to obtain $q_\theta$ which is then used as the trained
model.

\paragraph{Dataset.}
All of our experiments (as well as those in \newcite{hu2016harnessing}) use the SST2 dataset, a
binarized subset of
the popular Stanford Sentiment Treebank (SST)~\cite{socher2013recursive}.
The dataset includes phrase-level labels in addition to sentence-level labels (see~\tableref{sst2} for detailed statistics); following~\newcite{hu2016harnessing}, we use both types of labels for the comparisons in~\sectionref{huperform}. In all other experiments, we use only sentence-level labels, and our baseline model for all experiments is the CNN architecture from~\newcite{kim2014convolutional}. 
\begin{table}[b!]
\small
\begin{center}
\begin{tabular}{ lrrrr } 
 \toprule
Number of & Phrases & Train & Dev & Test\\ 
\midrule
Instances & 76961 & 6920 & 872 & 1821 \\
\textit{A-but-B} & 3.5\% & 11.1\% & 11.5\% & 11.5\% \\
Negations & 2.0\% & 17.5\% & 18.3\% & 17.2\% \\
Discourse & 5.0\% & 24.6\% & 26.0\% & 24.5\% \\
\bottomrule
\end{tabular}
\end{center}
\caption{Statistics of SST2 dataset. Here ``Discourse'' includes both \textit{A-but-B} and negation sentences. The mean length of sentences is in terms of the word count.\vspace{-1.5em}}
\label{tab:sst2}
\end{table}


\section{A Reanalysis}
\label{sec:hu}
In this section we reanalyze the effectiveness of the techniques of \newcite{hu2016harnessing} and find that most of the performance gain is due to projection and not knowledge distillation. The discrepancy with the original analysis can be attributed to the relatively small dataset and the resulting variance across random initializations. We start by analyzing the baseline CNN by~\newcite{kim2014convolutional} to point out the need for an averaged analysis.

\begin{figure}[ht!]
\includegraphics[scale=0.34]{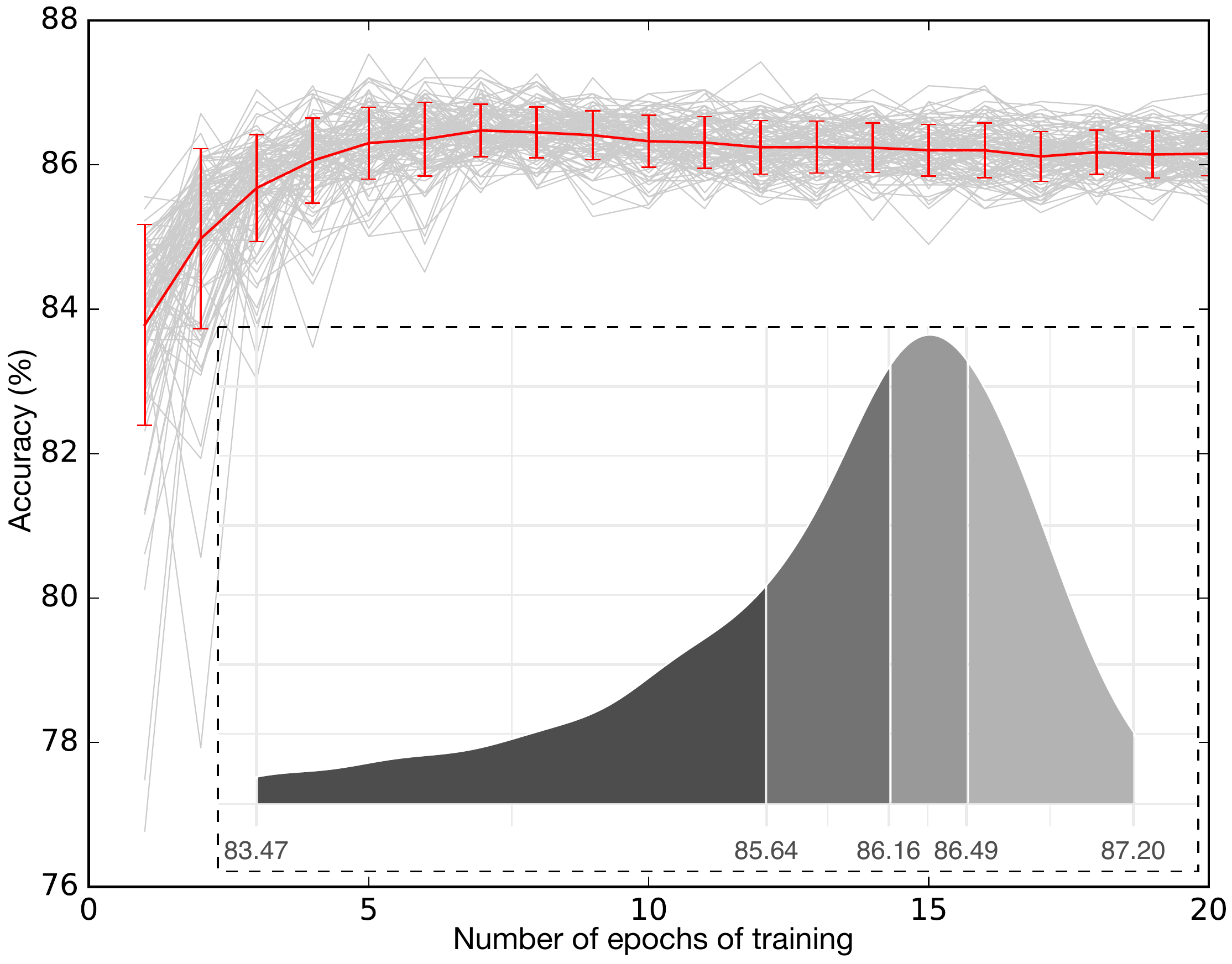}
\caption{Variation in models trained on SST-2 (sentence-only). Accuracies of 100
randomly initialized models are plotted against the number of epochs of training (in gray), along with their average accuracies (in red, with 95\% confidence interval error bars). The inset density plot shows the distribution
of accuracies when trained with early stopping.\vspace{-1.5em}}
\label{fig:variation}
\end{figure} 

\begin{figure*}[t!]
\centering
\includegraphics[scale=0.4]{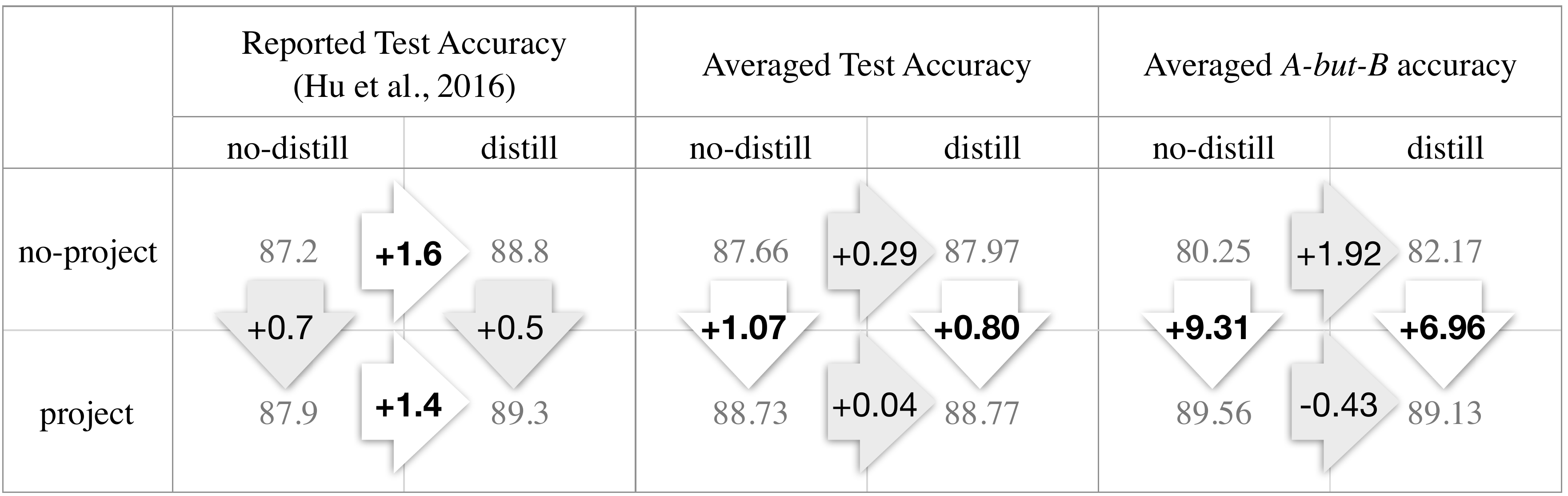}
\caption{Comparison of the accuracy improvements reported
in~\newcite{hu2016harnessing} and those obtained by averaging over 100 random
seeds. The last two columns show the (averaged) accuracy improvements for
\textit{A-but-B} style sentences.  All models use the publicly available
implementation of \newcite{hu2016harnessing} trained on phrase-level SST2 data.\vspace{-1.5em}
\label{fig:hu-performance}}
\end{figure*}

\subsection{Importance of Averaging}
We run the baseline CNN by~\newcite{kim2014convolutional} across 100 random seeds, training on sentence-level labels. We observe a large amount of variation from run-to-run, which is unsurprising given the small dataset size. The inset density plot in \figureref{variation} shows the range of accuracies (83.47 to 87.20) along with 25, 50 and 75 percentiles.\footnote{We use early stopping based on validation performance for all models in the density plot.} The figure also shows how the variance persists even after the average converges: the accuracies of 100 models trained for 20 epochs each are plotted in gray, and their average is shown in red.

We conclude that, to be reproducible, only averaged accuracies should be reported in this task and dataset. This mirrors the conclusion from a detailed analysis by~\newcite{reimers2017reporting} in the context of named entity recognition.

\subsection{Performance of~\newcite{hu2016harnessing}}
\label{sec:huperform}

We carry out an averaged analysis of the publicly available
implementation\footnote{\url{https://github.com/ZhitingHu/logicnn/}} of
~\newcite{hu2016harnessing}. Our analysis reveals that the reported performance
of their two mechanisms (projection and distillation) is in
fact affected by the high variability across random seeds. Our more robust averaged analysis yields a somewhat different conclusion of
their effectiveness.

In \figureref{hu-performance}, the first two columns show the reported
accuracies in \newcite{hu2016harnessing} for models trained with and without
distillation (corresponding to using values $\pi=1$ and
$\pi=0.95^t$ in the $t^\text{th}$ epoch, respectively). The two rows
show the results for models with and without a final projection into the
rule-regularized space.  We keep our hyper-parameters identical
to~\newcite{hu2016harnessing}.\footnote{In particular, $C = 6$ for projection.}

The baseline system (no-project, no-distill) is identical to the system of~\newcite{kim2014convolutional}. All the
systems are trained on the phrase-level SST2 dataset with early stopping on
the development set.
The number inside each arrow indicates the improvement in accuracy
by adding either the projection or the distillation component to the
training algorithm. Note that the reported figures suggest that while both
components help in improving accuracy, the distillation component is much
\emph{more} helpful than the projection component.

The next two columns, which show the results of repeating the
above analysis after averaging over 100 random seeds, contradict this claim. The averaged figures show lower overall accuracy increases, and, more importantly, they attribute these
improvements almost entirely to the projection component rather than the
distillation component. To confirm this result, we repeat our averaged analysis restricted to only \textit{``A-but-B''} sentences targeted by the rule (shown in the last two
columns). We again observe that
the effect of projection is pronounced, while distillation
offers little or no advantage in comparison.

\section{Contextualized Word Embeddings}
Traditional context-independent word embeddings like \texttt{word2vec}~\cite{mikolov2013efficient} or \texttt{GloVe}~\cite{pennington2014glove} are fixed vectors for every word in the vocabulary. In contrast, contextualized embeddings are dynamic representations, dependent on the current context of the word. We hypothesize that contextualized word embeddings might inherently capture these logic rules due to increasing the effective context size for the CNN layer in~\newcite{kim2014convolutional}. Following the recent success of ELMo~\cite{PetersELMo2018} in sentiment analysis, we utilize the TensorFlow Hub implementation of ELMo\footnote{\url{https://tfhub.dev/google/elmo/1}} and feed these contextualized embeddings into our CNN model. We fine-tune the ELMo LSTM weights along with the CNN weights on the downstream CNN task. As in \sectionref{hu}, we check performance with and without the final projection into the rule-regularized space. \\
We present our results in \tableref{elmo}. 

\begin{figure*}[t!]
\centering
\includegraphics[scale=0.37]{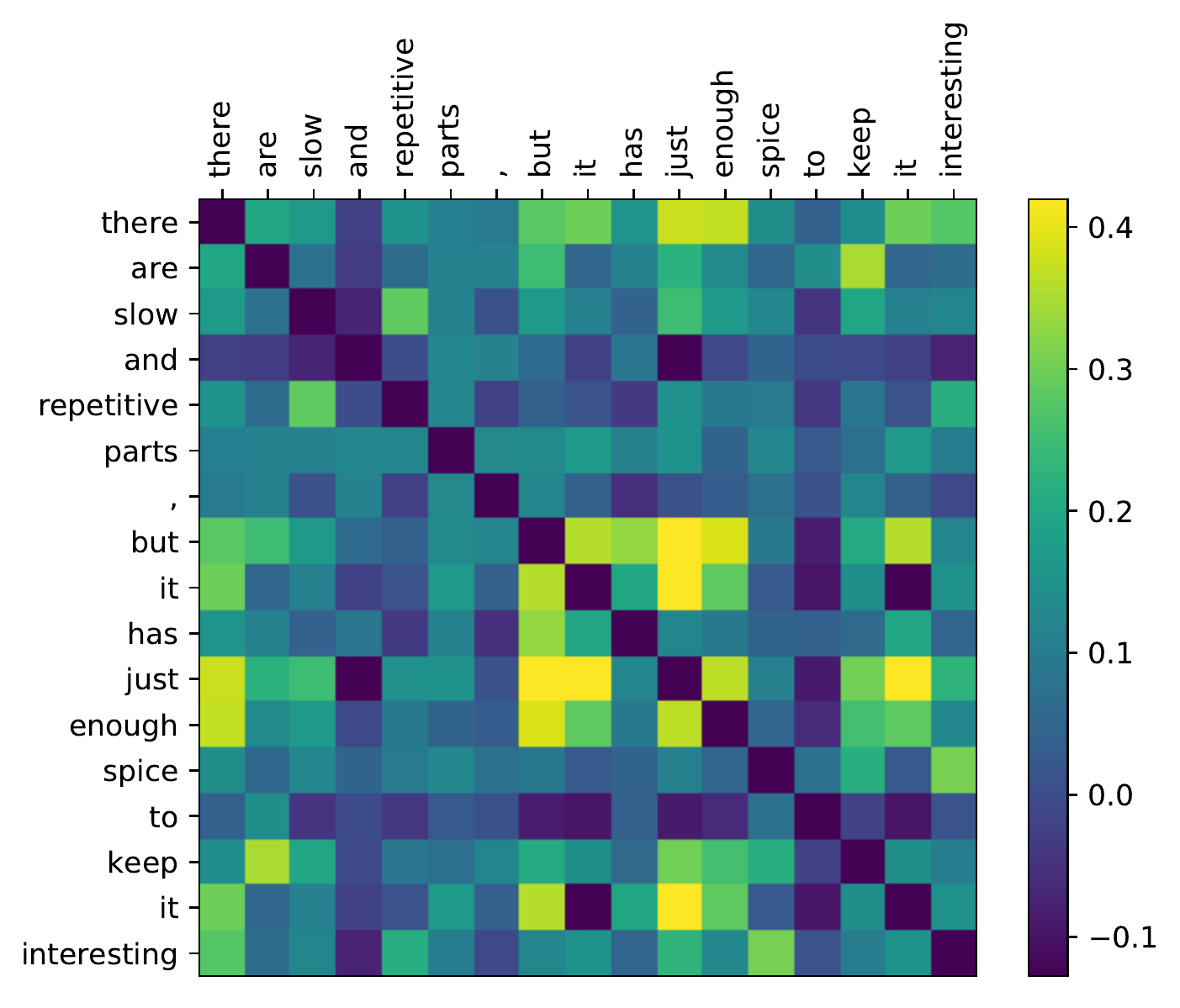}
\includegraphics[scale=0.37]{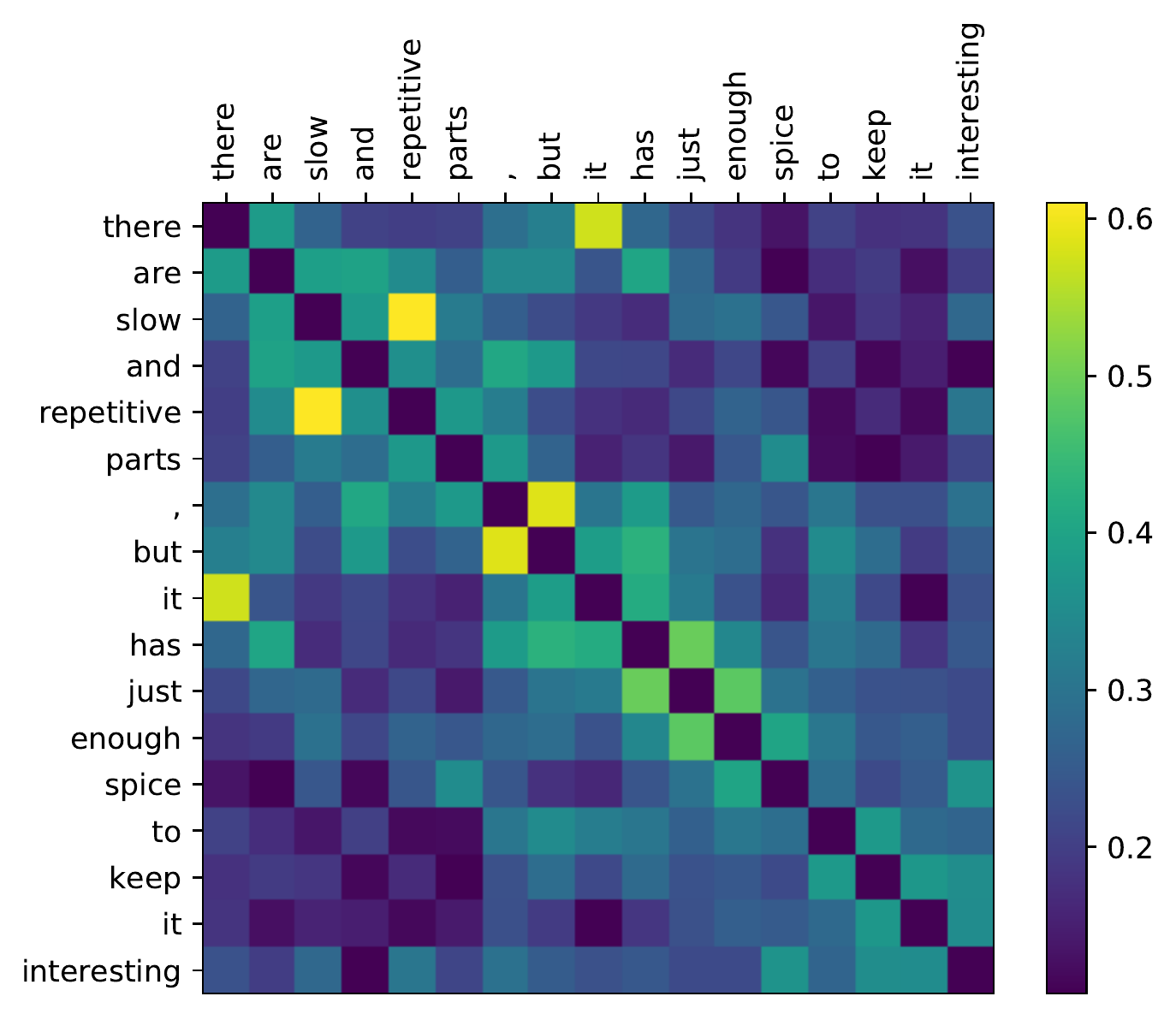}
\includegraphics[scale=0.37]{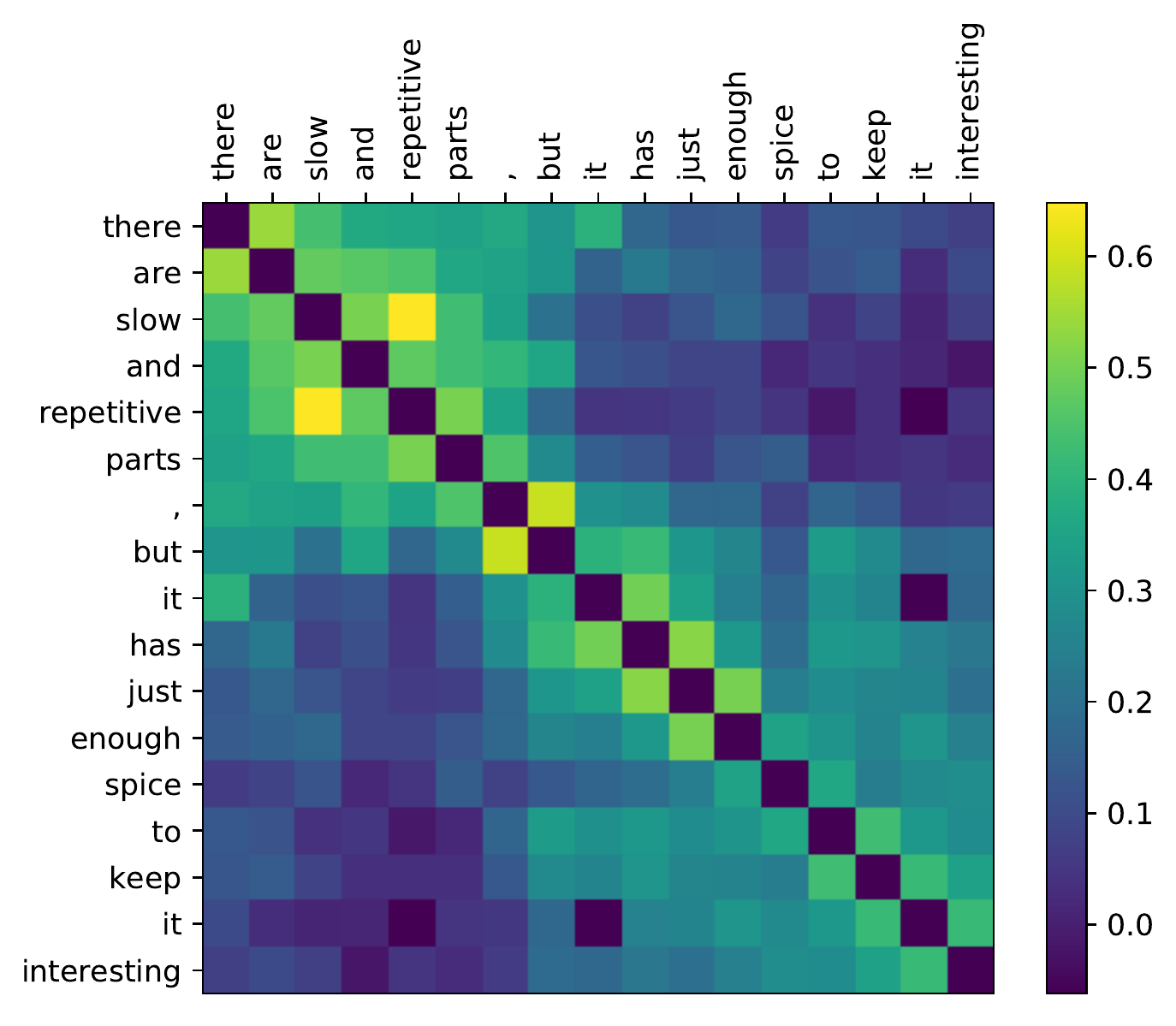}
\caption{Heat map showing the cosine similarity between pairs of word vectors within a single sentence. The left figure has fine-tuned \texttt{word2vec} embeddings. The middle figure contains the original ELMo embeddings without any fine-tuning. The right figure contains fine-tuned ELMo embeddings. For better visualization, the cosine similarity between identical words has been set equal to the minimum value in the heat map.}
\label{fig:visual}
\end{figure*}

Switching to ELMo word embeddings improves performance by 2.9 percentage points on an average, corresponding to about 53 test sentences. Of these, about 32 sentences (60\% of the improvement) correspond to \textit{A-but-B} and negation style sentences, which is substantial when considering that only 24.5\% of test sentences include these discourse relations (\tableref{sst2}). As further evidence that ELMo helps on these specific constructions, the non-ELMo baseline model (no-project, no-distill) gets 255 sentences wrong in the test corpus on average, only 89 (34.8\%) of  which are \textit{A-but-B} style or negations.
\paragraph{Statistical Significance:}
Using a two-sided Kolmogorov-Smirnov statistic~\cite{massey1951kolmogorov} with $\alpha = 0.001$ for the results in \tableref{elmo}, we find that ELMo and projection each yield statistically significant improvements, but distillation does not. Also, with ELMo, projection is not significant. Specific comparisons have been added in the Appendix, in \tableref{significance}.
\paragraph{KL Divergence Analysis:}
We observe no significant gains by projecting a trained ELMo model into an \textit{A-but-B} rule-regularized space, unlike the other models. We 
confirm that ELMo's predictions are much closer to the \textit{A-but-B} rule's manifold than those of the other models by computing $\mathrm{KL}(q_\theta||p_\theta)$ where $p_\theta$ and $q_\theta$ are the original and projected distributions: Averaged across all \textit{A-but-B} sentences and 100 seeds, this gives $0.27, 0.26$ and $0.13$ for the ~\newcite{kim2014convolutional}, ~\newcite{hu2016harnessing} with distillation and ELMo systems respectively.
\paragraph{Intra-sentence Similarity:} To understand the information captured by ELMo embeddings for \textit{A-but-B} sentences, we measure the cosine similarity between the word vectors of every pair of words within the \textit{A-but-B} sentence~\cite{peters2018dissecting}. We compare the intra-sentence similarity for fine-tuned \texttt{word2vec} embeddings (baseline), ELMo embeddings without fine-tuning and finally fine-tuned ELMo embeddings in \figureref{visual}. In the fine-tuned ELMo embeddings, we notice the words within the \textit{A} and within the \textit{B} part of the \textit{A-but-B} sentence share the same part of the vector space. This pattern is less visible in the ELMo embeddings without fine-tuning and absent in the \texttt{word2vec} embeddings. This observation is indicative of ELMo's ability to learn specific rules for  \textit{A-but-B} sentences in sentiment classification. More intra-sentence similarity heatmaps for \textit{A-but-B} sentences are in \figureref{appendvisual}.

\begin{table}
\small
\begin{center}
\begin{tabular}{ rrccc } 
 \toprule
 \multicolumn{2}{c}{Model} & Test & \textit{but}  & \textit{but} or \textit{neg}\\ 
\midrule
no-distill & no-project & 85.98 & 78.69 & 80.13 \\
no-distill & project & 86.54 & 83.40 & - \\
 \midrule
 distill \tablefootnote{Trained on sentences and not phrase-level labels for a fair comparison with baseline and ELMo, unlike \sectionref{huperform}.} & no-project & 86.11 & 79.04 & - \\
distill & project & 86.62 & 83.32  & - \\
 \midrule
 ELMo & no-project & 88.89 & 86.51 & 87.24 \\
 ELMo & project & 88.96 & 87.20 & - \\
 \bottomrule
\end{tabular}
\end{center}
\caption{Average performance (across 100 seeds) of ELMo on the SST2 task. We show performance on \textit{A-but-B} sentences (``\textit{but}''), negations (``\textit{neg}'')\vspace{-1.5em}.}
\label{tab:elmo}
\end{table}

\section{Crowdsourced Experiments}

We conduct a crowdsourced analysis that reveals that SST2 data has significant levels of ambiguity even for human labelers. We discover that ELMo's performance improvements over the baseline are robust across varying levels of ambiguity, whereas the advantage of \newcite{hu2016harnessing} is reversed in sentences of low ambiguity (restricting to \textit{A-but-B} style sentences).


Our crowdsourced experiment was conducted on Figure Eight.\footnote{ \url{https://www.figure-eight.com/}} \emph{Nine} workers scored the sentiment of each \textit{A-but-B} and negation sentence in the test SST2 split as 0 (negative), 0.5 (neutral) or 1 (positive).  (SST originally had \emph{three} crowdworkers choose a sentiment rating from 1 to 25 for every phrase.) More details regarding the crowd experiment's parameters have been provided in \appendixref{appcrowd}.

We average the scores across all users for each sentence. Sentences with a score in the range $(x, 1]$ are marked as positive (where $x\in[0.5,1)$), sentences in $[0, 1-x)$ marked as negative, and sentences in $[1-x, x]$ are marked as neutral. For instance, ``{\em flat , but with a revelatory performance by michelle williams}'' (score=0.56) is neutral when $x=0.6$.\footnote{More examples of neutral sentences have been provided in the Appendix in \tableref{neutral}, as well as a few ``flipped'' sentences receiving an average score opposite to their SST2 label (\tableref{flipped}).}  We present statistics of our dataset\footnote{The dataset along with source code can be found in \url{https://github.com/martiansideofthemoon/logic-rules-sentiment}.} in \tableref{crowd_all}. Inter-annotator agreement was computed using Fleiss' Kappa ($\kappa$). As expected, inter-annotator agreement is higher for higher thresholds (less ambiguous sentences). According to~\newcite{landis1977measurement}, $\kappa \in (0.2, 0.4]$ corresponds to ``fair agreement'', whereas $\kappa \in (0.4, 0.6]$ corresponds to ``moderate agreement''.

\begin{figure}[ht!]
\centering
\includegraphics[width=\linewidth]{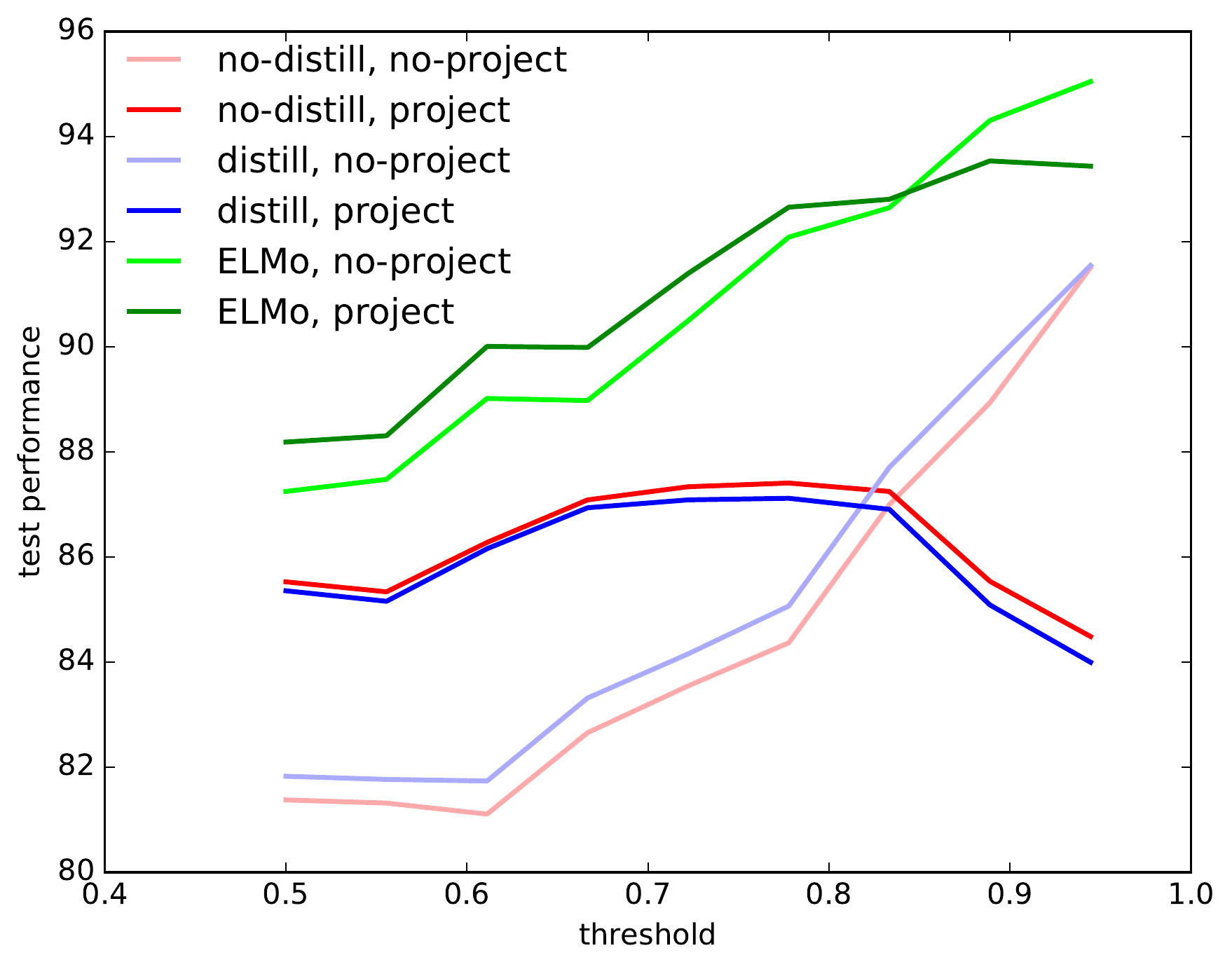}
\caption{Average performance on the \textit{A-but-B}  part of the crowd-sourced dataset (210 sentences, 100 seeds)). For each threshold, only non-neutral sentences are used for evaluation.\vspace{-1.5em}}
\label{fig:crowd}
\end{figure}

We next compute the accuracy of our model for each threshold by removing the corresponding neutral sentences. Higher thresholds correspond to sets of less ambiguous sentences.
\tableref{crowd_all} shows that ELMo's performance gains in \tableref{elmo} extends across all thresholds. In \figureref{crowd} we compare all the models on the \textit{A-but-B} sentences in this set. Across all thresholds, we notice trends similar to previous sections: 1) ELMo performs the best among all models on \textit{A-but-B} style sentences, and projection results in only a slight improvement; 2) models in~\newcite{hu2016harnessing} (with and without distillation) benefit considerably from projection; but 3) distillation offers little improvement (with or without projection). Also, as the ambiguity threshold increases, we see decreasing gains from projection on all models. In fact, beyond the 0.85 threshold, projection degrades the average performance, indicating that projection is useful for more ambiguous sentences.
\begin{table}
\small
\begin{center}
\begin{tabular}{ rcccc } 
 \toprule
Threshold  & 0.50 & 0.66 & 0.75 & 0.90  \\
 \midrule
Neutral Sentiment & 10 & 70 & 95 & 234 \\
Flipped Sentiment  & 15 & 4 & 2 & 0 \\
Fleiss' Kappa ($\kappa$) & 0.38 & 0.42 & 0.44 & 0.58 \\
\midrule
no-distill, no-project & 81.32 & 83.54 & 84.54 & 87.55\\
ELMo, no-project & 87.56 & 90.00 & 91.31 & 93.14\\
\bottomrule
\end{tabular}
\end{center}
\caption{ Number of sentences in the crowdsourced study (447 sentences) which got marked as neutral and which got the opposite of their labels in the SST2 dataset, using various thresholds. Inter-annotator agreement is computed using Fleiss' Kappa. Average accuracies of the baseline and ELMo (over 100 seeds) on  non-neutral sentences are also shown.\vspace{-1.5em}}
\label{tab:crowd_all}
\end{table}


\section{Conclusion}
We present an analysis comparing techniques for incorporating logic rules into sentiment classification systems. Our analysis included a meta-study highlighting the issue of stochasticity in performance across runs and the inherent ambiguity in the sentiment classification task itself, which was tackled using an averaged analysis and a crowdsourced experiment identifying ambiguous sentences. We present evidence that a recently proposed contextualized word embedding model (ELMo)~\cite{PetersELMo2018} implicitly learns logic rules for sentiment classification of complex sentences like \textit{A-but-B} sentences. Future work includes a fine-grained quantitative study of ELMo word vectors  for logically complex sentences along the lines of~\newcite{peters2018dissecting}.

\bibliography{emnlp2018}

\begin{thebibliography}{17}
\expandafter\ifx\csname natexlab\endcsname\relax\def\natexlab#1{#1}\fi

\bibitem[{Ganchev et~al.(2010)Ganchev, Gillenwater, Taskar
  et~al.}]{ganchev2010posterior}
Kuzman Ganchev, Jennifer Gillenwater, Ben Taskar, et~al. 2010.
\newblock Posterior regularization for structured latent variable models.
\newblock \emph{Journal of Machine Learning Research}, 11(Jul):2001--2049.

\bibitem[{Hinton et~al.(2015)Hinton, Vinyals, and Dean}]{hinton2015distilling}
Geoffrey Hinton, Oriol Vinyals, and Jeff Dean. 2015.
\newblock Distilling the knowledge in a neural network.
\newblock \emph{NIPS Deep Learning and Representation Learning Workshop}.

\bibitem[{Hu et~al.(2016)Hu, Ma, Liu, Hovy, and Xing}]{hu2016harnessing}
Zhiting Hu, Xuezhe Ma, Zhengzhong Liu, Eduard Hovy, and Eric Xing. 2016.
\newblock Harnessing deep neural networks with logic rules.
\newblock In \emph{Association for Computational Linguistics (ACL)}.

\bibitem[{Iyyer et~al.(2015)Iyyer, Manjunatha, Boyd-Graber, and
  Daum{\'e}~III}]{iyyer2015deep}
Mohit Iyyer, Varun Manjunatha, Jordan Boyd-Graber, and Hal Daum{\'e}~III. 2015.
\newblock Deep unordered composition rivals syntactic methods for text
  classification.
\newblock In \emph{Association for Computational Linguistics (ACL)}.

\bibitem[{Kim(2014)}]{kim2014convolutional}
Yoon Kim. 2014.
\newblock Convolutional neural networks for sentence classification.
\newblock In \emph{Empirical Methods in Natural Language Processing (EMNLP)}.

\bibitem[{Landis and Koch(1977)}]{landis1977measurement}
J~Richard Landis and Gary~G Koch. 1977.
\newblock The measurement of observer agreement for categorical data.
\newblock \emph{Biometrics}, pages 159--174.

\bibitem[{Li et~al.(2017)Li, Zhao, Liu, Hu, and Du}]{li2017initializing}
Shen Li, Zhe Zhao, Tao Liu, Renfen Hu, and Xiaoyong Du. 2017.
\newblock Initializing convolutional filters with semantic features for text
  classification.
\newblock In \emph{Empirical Methods in Natural Language Processing (EMNLP)}.

\bibitem[{Massey~Jr(1951)}]{massey1951kolmogorov}
Frank~J Massey~Jr. 1951.
\newblock The {K}olmogorov-{S}mirnov test for goodness of fit.
\newblock \emph{Journal of the American statistical Association},
  46(253):68--78.

\bibitem[{Mikolov et~al.(2013)Mikolov, Chen, Corrado, and
  Dean}]{mikolov2013efficient}
Tomas Mikolov, Kai Chen, Greg Corrado, and Jeffrey Dean. 2013.
\newblock Efficient estimation of word representations in vector space.
\newblock \emph{arXiv preprint arXiv:1301.3781}.

\bibitem[{Pang and Lee(2005)}]{pang2005seeing}
Bo~Pang and Lillian Lee. 2005.
\newblock Seeing stars: Exploiting class relationships for sentiment
  categorization with respect to rating scales.
\newblock In \emph{Association for Computational Linguistics (ACL)}.

\bibitem[{Pennington et~al.(2014)Pennington, Socher, and
  Manning}]{pennington2014glove}
Jeffrey Pennington, Richard Socher, and Christopher Manning. 2014.
\newblock {GloVe}: Global vectors for word representation.
\newblock In \emph{Empirical Methods in Natural Language Processing (EMNLP)}.

\bibitem[{Peters et~al.(2018{\natexlab{a}})Peters, Neumann, Iyyer, Gardner,
  Clark, Lee, and Zettlemoyer.}]{PetersELMo2018}
Matthew~E. Peters, Mark Neumann, Mohit Iyyer, Matt Gardner, Christopher Clark,
  Kenton Lee, and Luke Zettlemoyer. 2018{\natexlab{a}}.
\newblock Deep contextualized word representations.
\newblock In \emph{North American Association for Computational Linguistics
  (NAACL)}.

\bibitem[{Peters et~al.(2018{\natexlab{b}})Peters, Neumann, tau Yih, and
  Zettlemoyer}]{peters2018dissecting}
Matthew~E. Peters, Mark Neumann, Wen tau Yih, and Luke Zettlemoyer.
  2018{\natexlab{b}}.
\newblock Dissecting contextual word embeddings: Architecture and
  representation.
\newblock In \emph{Empirical Methods in Natural Language Processing (EMNLP)}.

\bibitem[{Reimers and Gurevych(2017)}]{reimers2017reporting}
Nils Reimers and Iryna Gurevych. 2017.
\newblock Reporting score distributions makes a difference: Performance study
  of {LSTM}-networks for sequence tagging.
\newblock In \emph{Empirical Methods in Natural Language Processing (EMNLP)}.

\bibitem[{Socher et~al.(2013)Socher, Perelygin, Wu, Chuang, Manning, Ng, and
  Potts}]{socher2013recursive}
Richard Socher, Alex Perelygin, Jean Wu, Jason Chuang, Christopher~D Manning,
  Andrew Ng, and Christopher Potts. 2013.
\newblock Recursive deep models for semantic compositionality over a sentiment
  treebank.
\newblock In \emph{Empirical Methods in Natural Language Processing (EMNLP)}.

\bibitem[{Tai et~al.(2015)Tai, Socher, and Manning}]{tai2015improved}
Kai~Sheng Tai, Richard Socher, and Christopher~D Manning. 2015.
\newblock Improved semantic representations from tree-structured long
  short-term memory networks.
\newblock In \emph{Association for Computational Linguistics (ACL)}.

\bibitem[{Wang and Manning(2012)}]{wang2012simple}
S.~I. Wang and C.~Manning. 2012.
\newblock Baselines and bigrams: Simple, good sentiment and text
  classification.
\newblock In \emph{Association for Computational Linguistics (ACL)}.

\end{thebibliography}
\bibliographystyle{acl_natbib_nourl}

\newpage
\appendix

\section*{Appendix}

\setcounter{table}{0} \renewcommand{\thetable}{A\arabic{table}} 
\setcounter{figure}{0} \renewcommand{\thefigure}{A\arabic{figure}}

\section{Crowdsourcing Details}
\label{appendix:appcrowd}
Crowd workers residing in five English-speaking countries (United States, United Kingdom, New Zealand, Australia and Canada) were hired. Each crowd worker had a Level 2 or higher rating on Figure Eight, which corresponds to a ``group of more experienced, higher accuracy contributors''. Each contributor had to pass a test questionnaire to be eligible to take part in the experiment. Test questions were also hidden throughout the task and untrusted contributions were removed from the final dataset. For greater quality control, an upper limit of 75 judgments per contributor was enforced.\\
Crowd workers were paid a total of \$1 for  50 judgments. An internal unpaid workforce (including the first and second author of the paper) of 7 contributors was used to speed up data collection.

\begin{table*}
\begin{center}
\begin{tabular}{ ccccm{8cm} } 
 \toprule
 \multicolumn{3}{c}{\# Judgments} & Average & Sentence  \\
 Positive & Negative & Neutral & & \\
\midrule
 1 & 1 & 7 & 0.50 & the fight scenes are fun , but it grows tedious \\
 \midrule
 3 & 2 & 4 & 0.56 & it 's not exactly a gourmet meal but the fare is fair , even coming from the drive thru \\
 \midrule
 2 & 3 & 4 & 0.44 & propelled not by characters but by caricatures \\
 \midrule
 4 & 2 & 3 & 0.61 & not everything works , but the average is higher than in mary and most other recent comedies \\
 \bottomrule
\end{tabular}
\end{center}
\caption{Examples of neutral sentences for a threshold of 0.66}
\label{tab:neutral}
\end{table*}

\begin{table*}
\begin{center}
\begin{tabular}{ cccccm{7cm} } 
 \toprule
 \multicolumn{3}{c}{\# Judgments} & Average & Original & Sentence  \\
 Positive & Negative & Neutral & & \\
\midrule
1 & 5 & 3 & 0.28 & Positive & de niro and mcdormand give solid performances , but their screen time is sabotaged by the story 's inability to create interest \\
\midrule
6 & 0 & 3 & 0.83 & Negative & son of the bride may be a good half hour too long but comes replete with a flattering sense of mystery and quietness \\
\midrule
0 & 5 & 4 & 0.22 & Positive & wasabi is slight fare indeed , with the entire project having the feel of something tossed off quickly ( like one of hubert 's punches ) , but it should go down smoothly enough with popcorn \\
 \bottomrule
\end{tabular}
\end{center}
\caption{Examples of flipped sentiment sentences, for a threshold of 0.66}
\label{tab:flipped}
\end{table*}

\begin{table*}
\begin{center}
\begin{tabular}{ rrcrrc } 
 \toprule
 \multicolumn{2}{c}{Model 1} & vs & \multicolumn{2}{c}{Model 2} & Significant \\
 \midrule
 distill & no-project & & distill & project & Yes \\
  no-distill & no-project & & no-distill & project & Yes \\
  ELMo & no-project & & ELMo & project & No \\
 \midrule
 no-distill & no-project & & distill & no-project & No \\
 no-distill & project & & distill & project & No \\
 \midrule
 no-distill & no-project & & ELMo & no-project & Yes \\
 distill & no-project & & ELMo & no-project & Yes \\
  no-distill & project & & ELMo & project & Yes \\
 distill & project & & ELMo & project & Yes \\
 \bottomrule
\end{tabular}
\end{center}
\caption{Statistical significance using a two-sided Kolmogorov-Smirnov statistic~\cite{massey1951kolmogorov} with $\alpha = 0.001$.}
\label{tab:significance}
\end{table*}

\begin{figure*}[t!]
\centering
\includegraphics[scale=0.37]{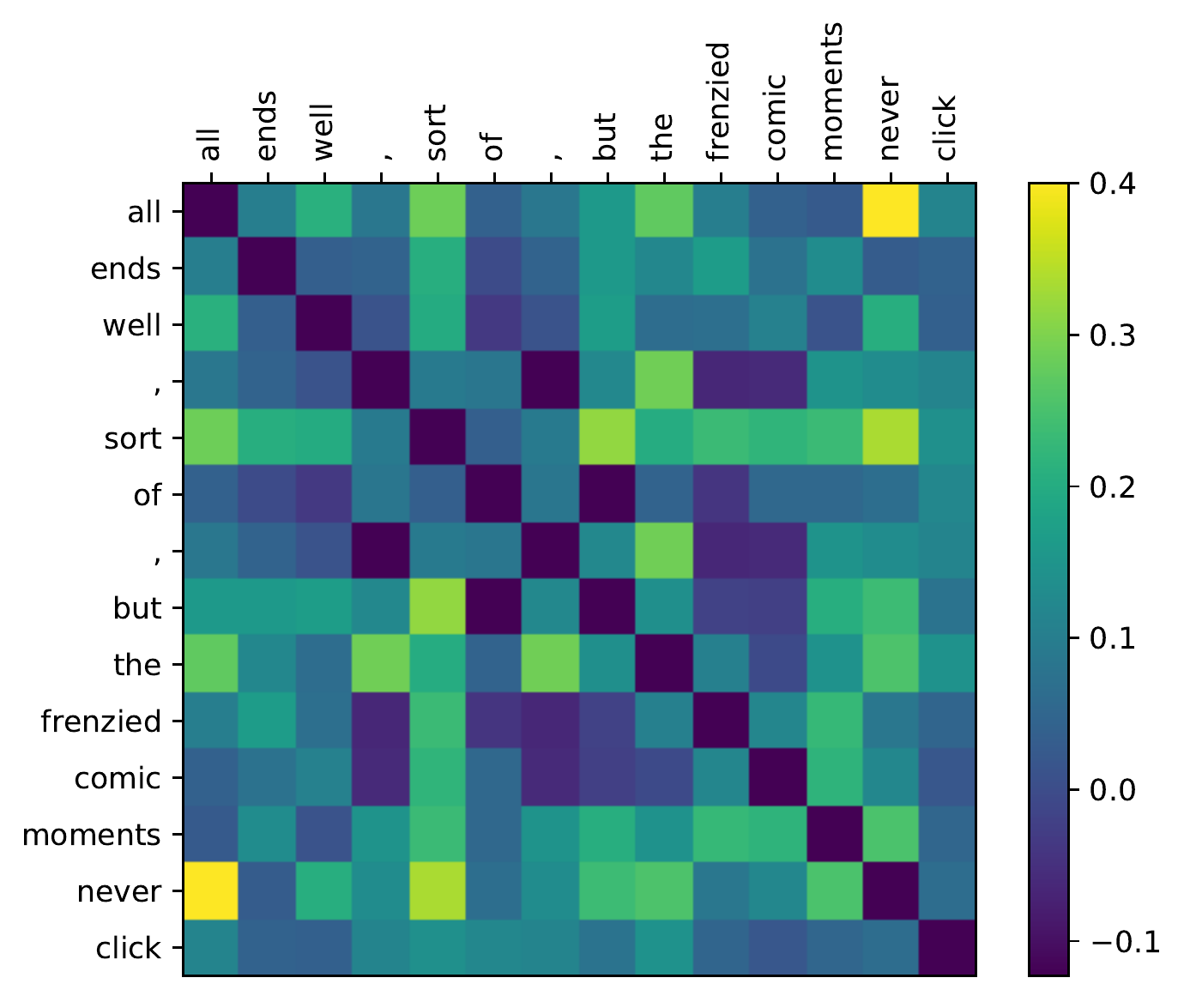}
\includegraphics[scale=0.37]{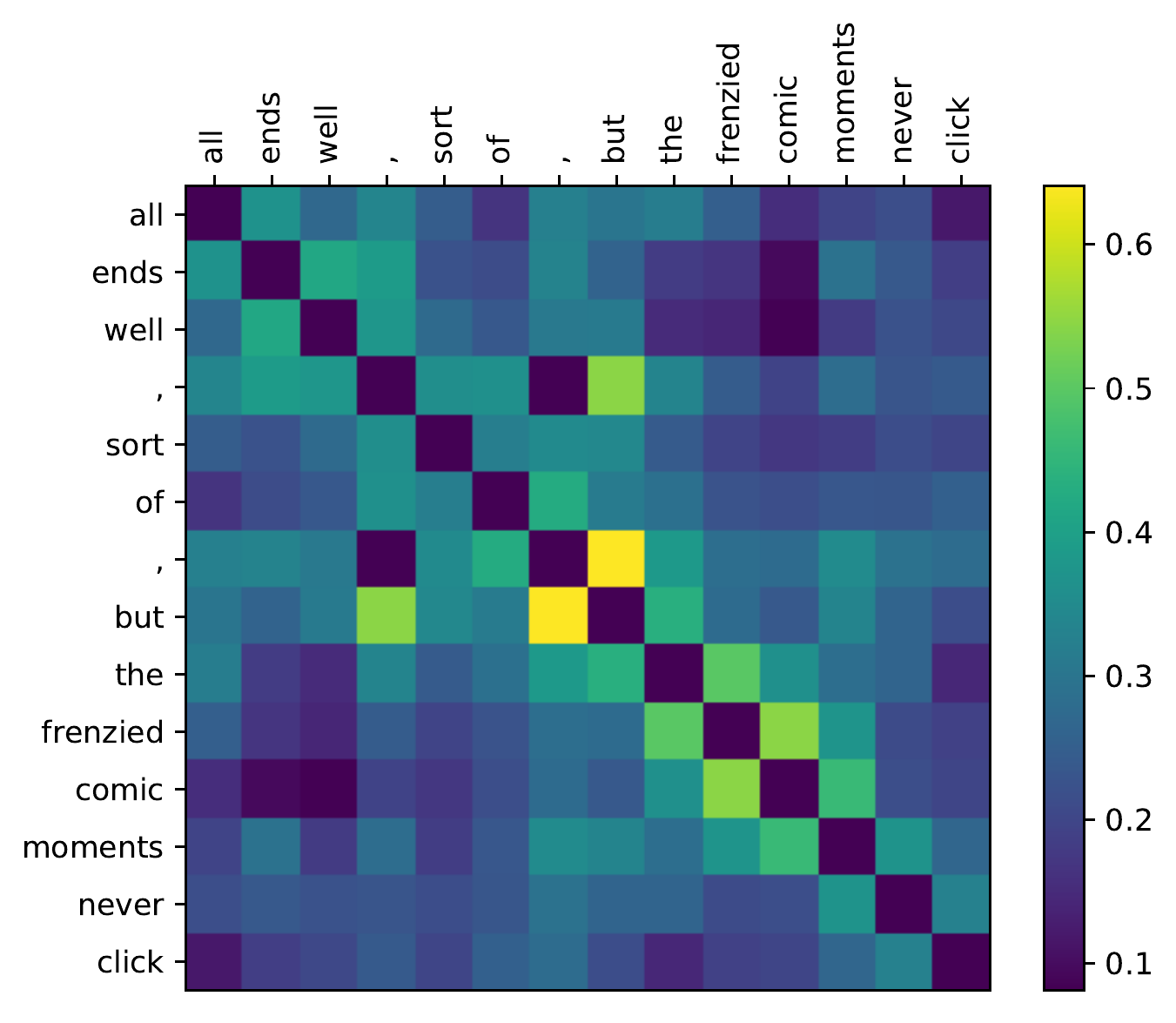}
\includegraphics[scale=0.37]{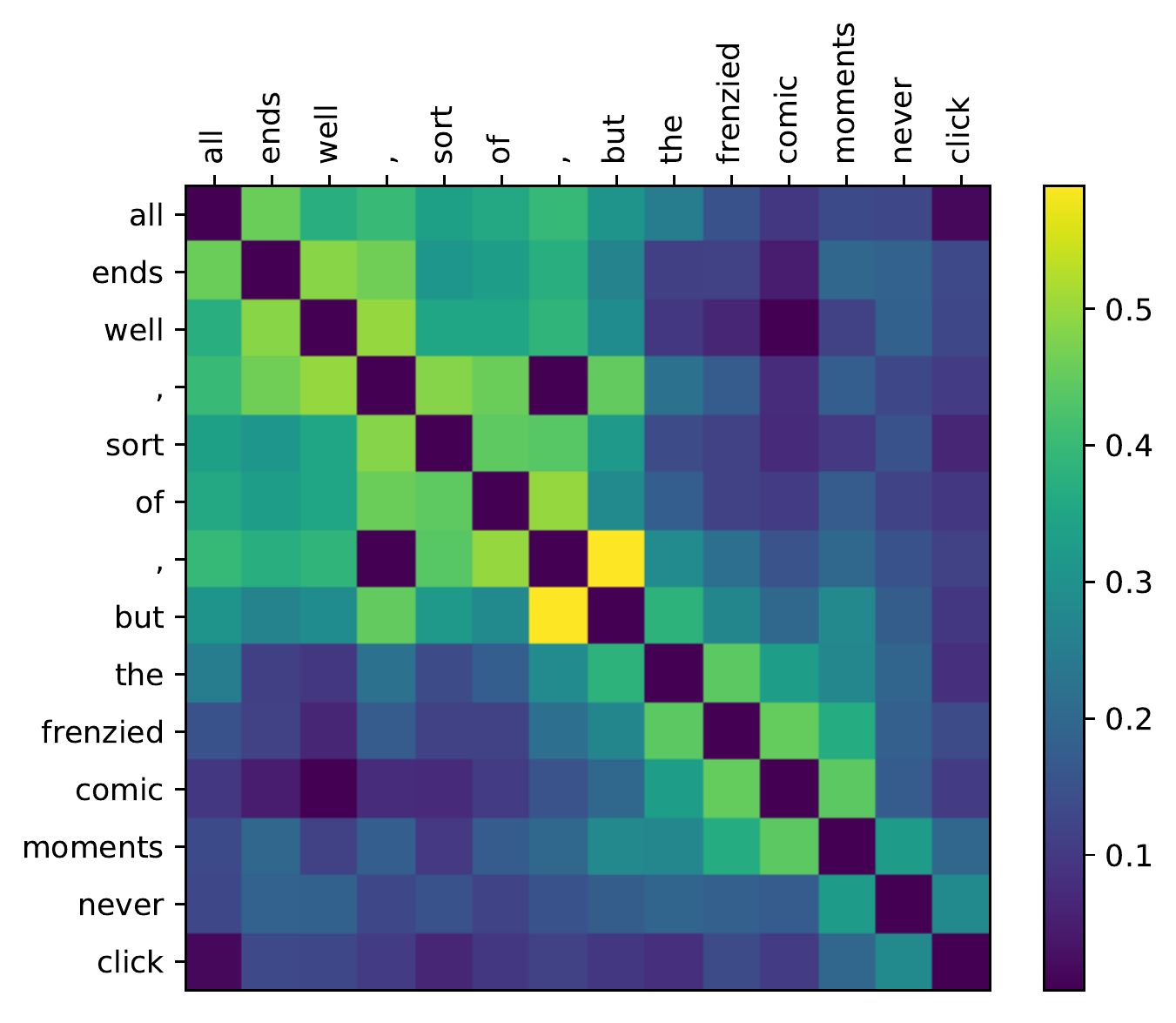}
\includegraphics[scale=0.37]{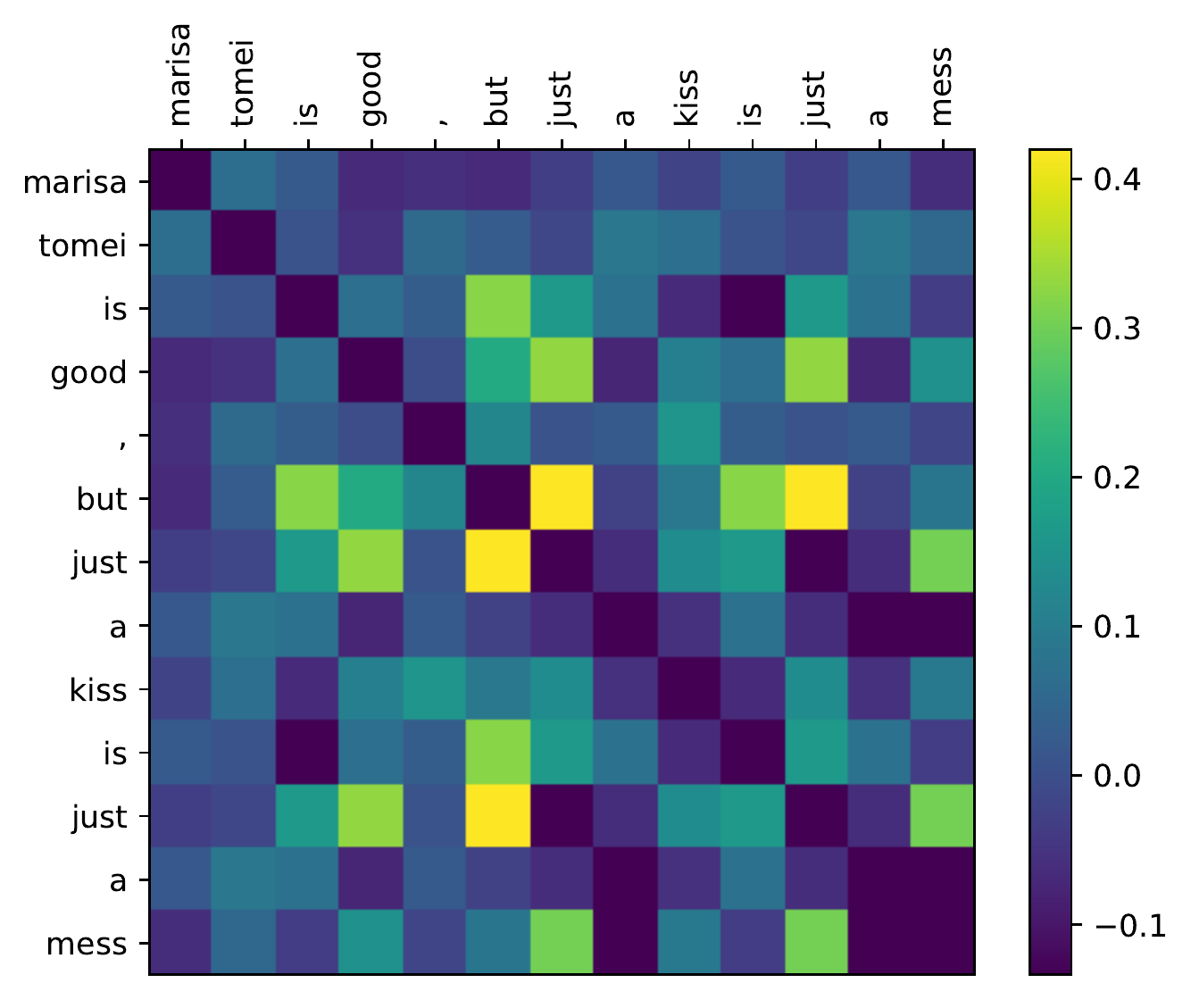}
\includegraphics[scale=0.37]{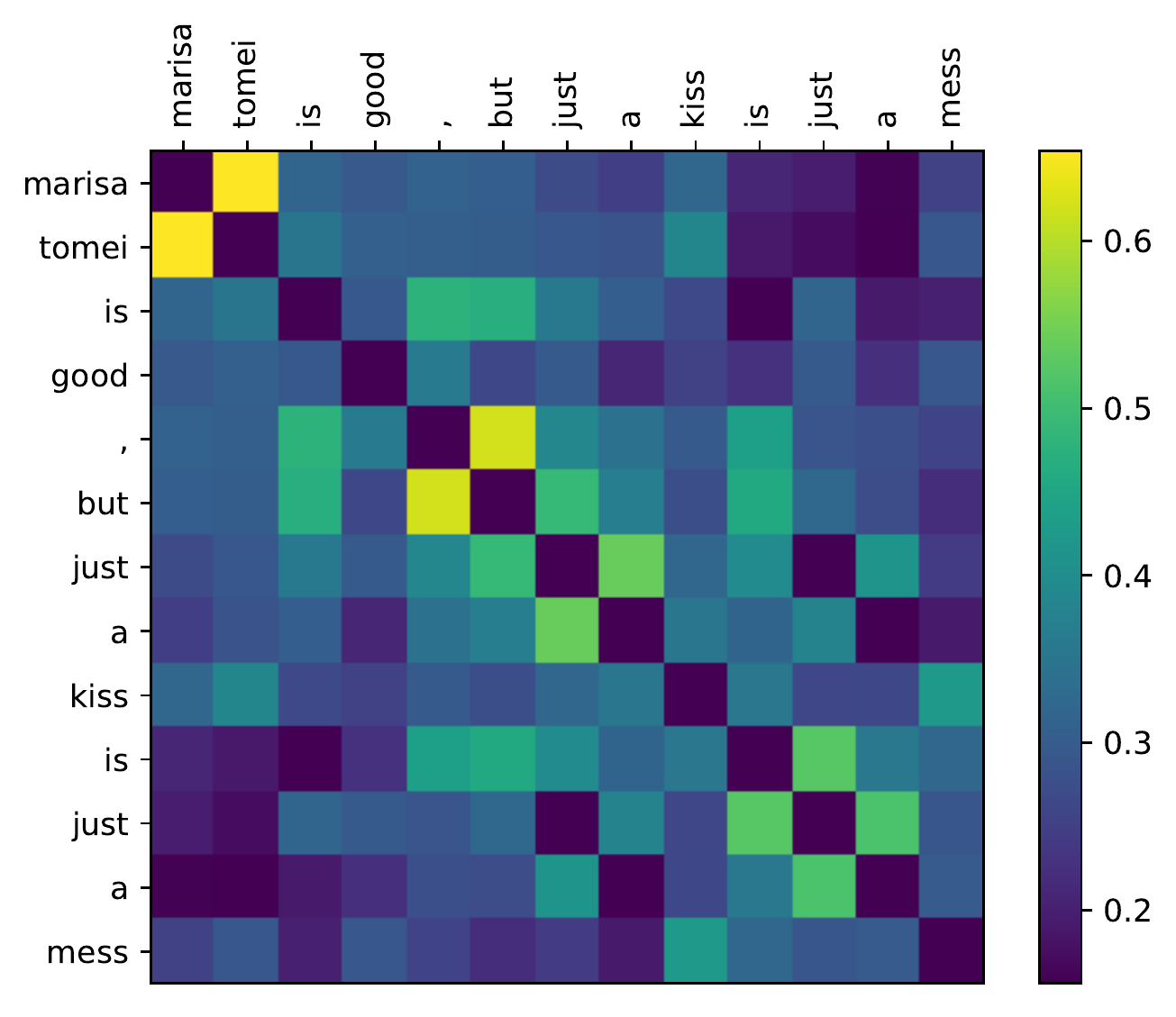}
\includegraphics[scale=0.37]{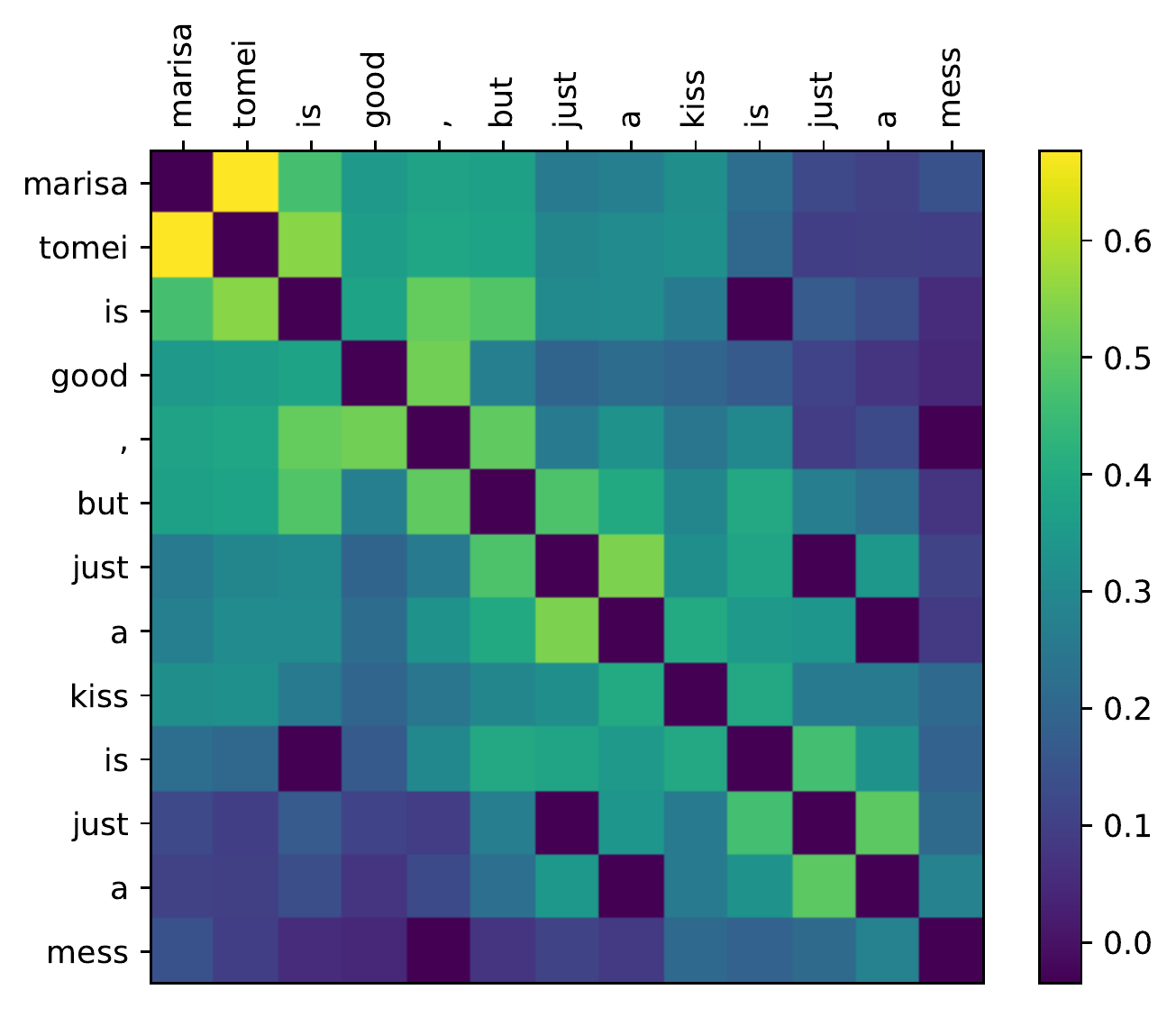}
\includegraphics[scale=0.36]{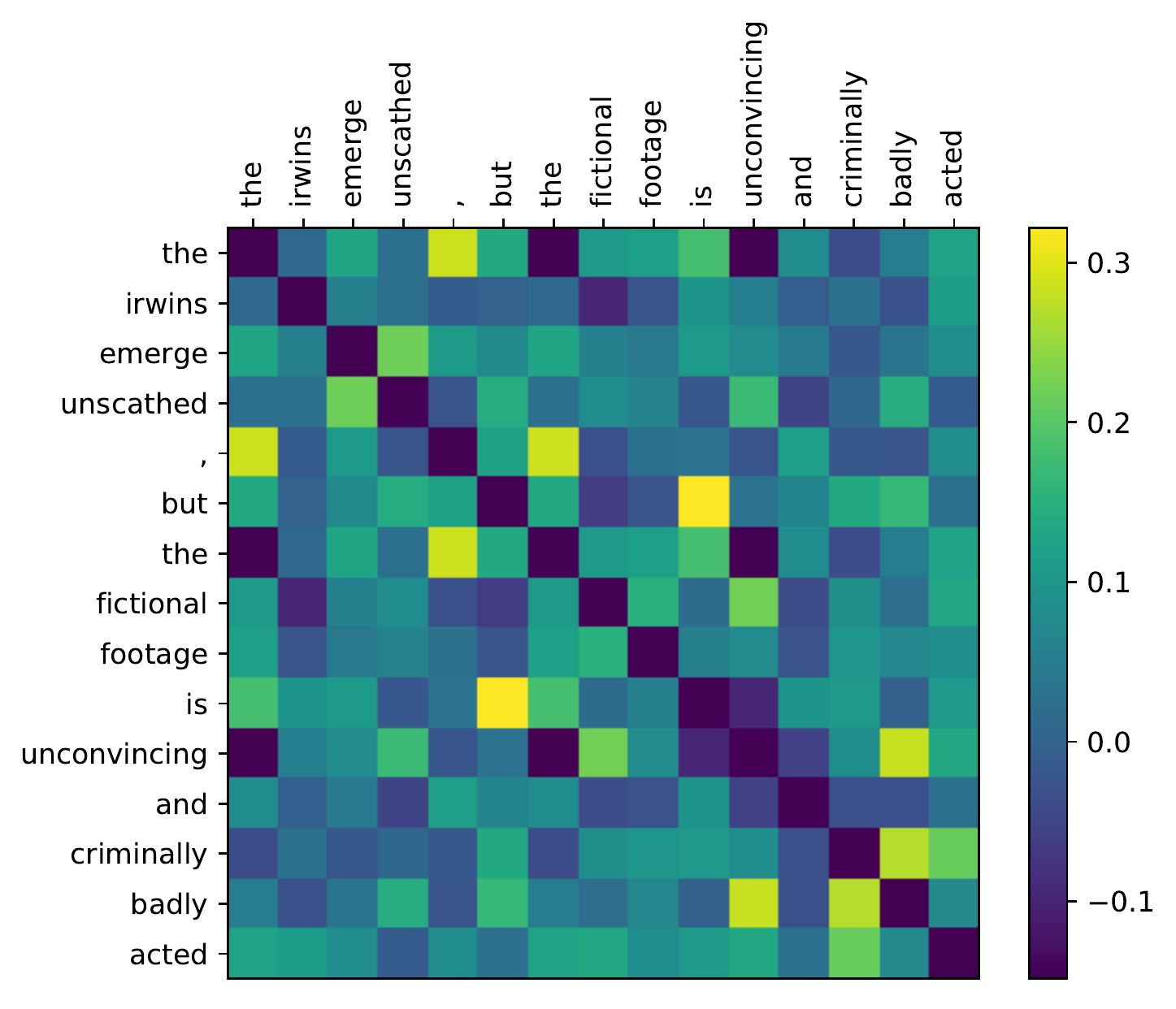}
\includegraphics[scale=0.36]{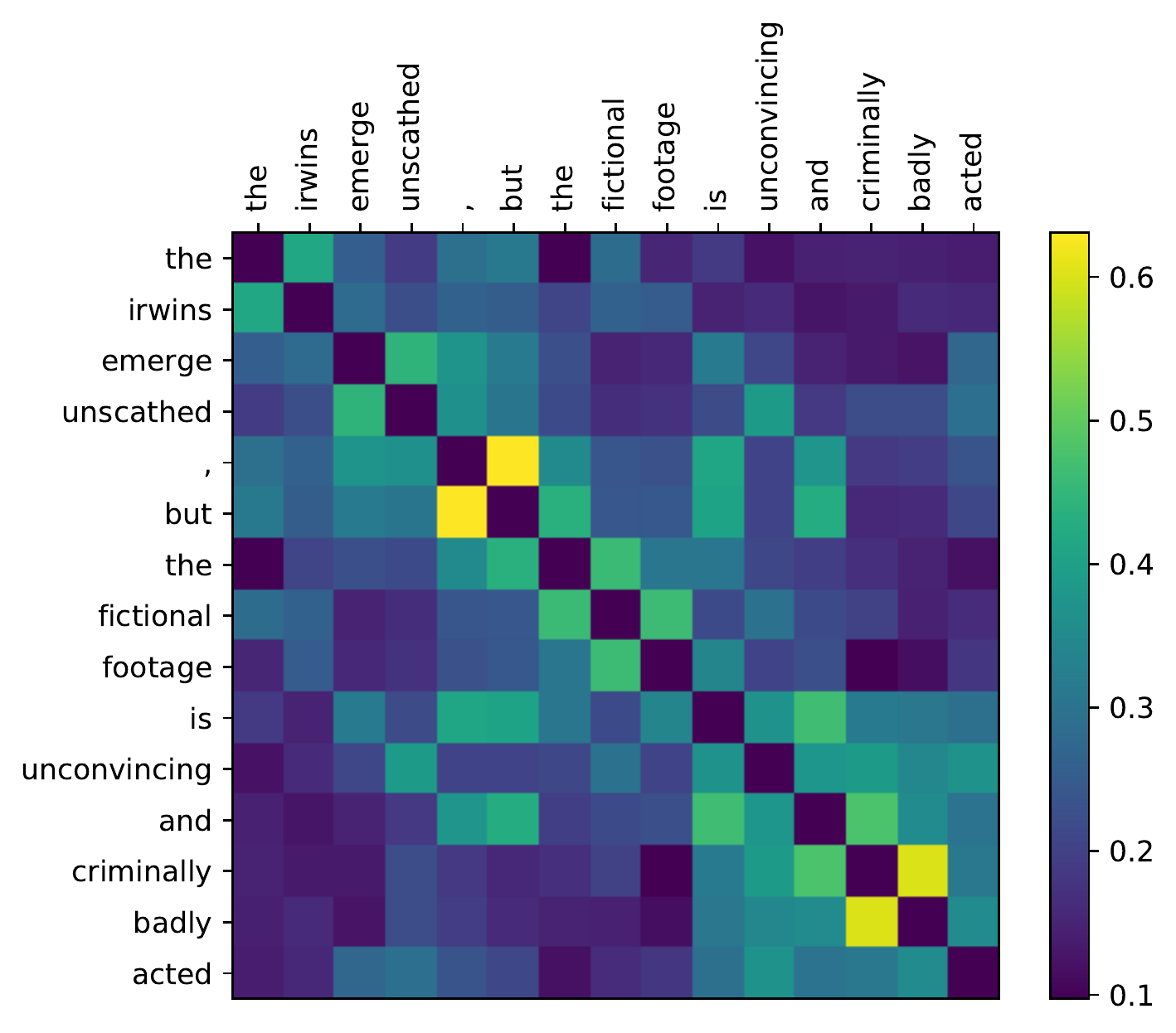}
\includegraphics[scale=0.36]{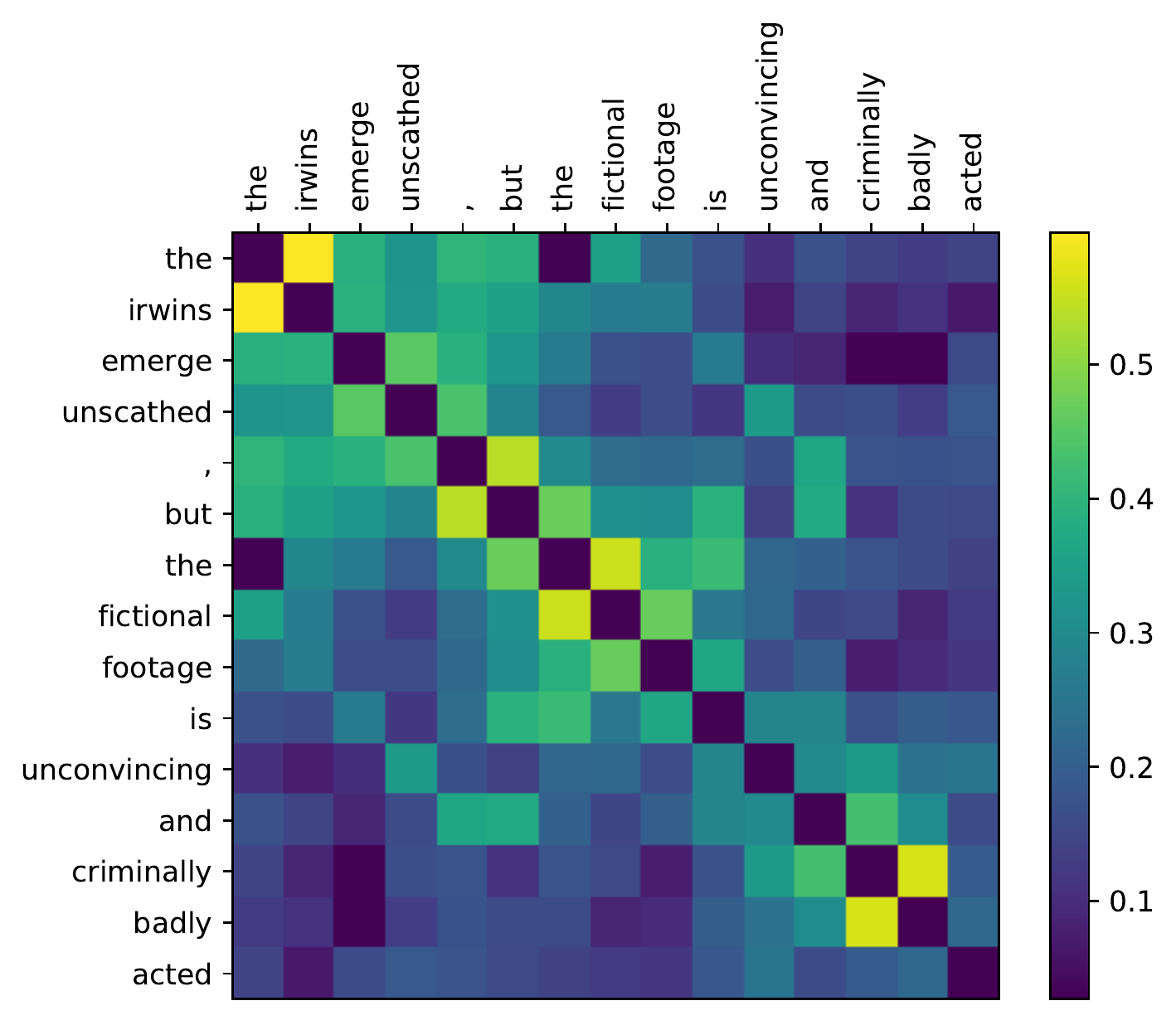}
\caption{Heat map showing the cosine similarity between pairs of word vectors within a single sentence. The leftmost column has \texttt{word2vec}~\cite{mikolov2013efficient} embeddings, fine-tuned on the downstream task (SST2). The middle column contains the original ELMo embeddings~\cite{PetersELMo2018} without any fine-tuning. The representations from the three layers (token layer and two LSTM layers) have been averaged. The rightmost column contains ELMo embeddings fine-tuned on the downstream task. For better visualization, the cosine similarity between identical words has been set equal to the minimum value in the map.}
\label{fig:appendvisual}
\end{figure*}

\end{document}